\documentclass[letterpaper, 10 pt, conference]{ieeeconf}  % Comment this line out if you need a4paper

\IEEEoverridecommandlockouts                              % This command is only needed if 
\usepackage{times}
\usepackage{epsfig}
\usepackage{graphicx}
\usepackage{amsmath}
\usepackage{amssymb}
\usepackage{verbatim}
\usepackage{mathtools}
\let\labelindent\relax
\usepackage{enumitem}
\usepackage{booktabs}
\usepackage{xcolor}
\usepackage{multirow}
\usepackage{subcaption}
\usepackage{algorithm} 
\usepackage{algpseudocode} 
\usepackage[font={small}]{caption}
\usepackage[pagebackref=true,breaklinks=true,letterpaper=true,colorlinks,citecolor=blue,linkcolor=blue,bookmarks=false]{hyperref}

\title{\LARGE \bf
SKT-Hang: Hanging Everyday Objects via Object-Agnostic Semantic Keypoint Trajectory Generation
}

\author{Chia-Liang Kuo$^{1}$ \, Yu-Wei Chao$^{2}$ \, Yi-Ting Chen$^{1}$ \\ $^{1}$ National Yang Ming Chiao Tung University \, $^{2}$ NVIDIA}
\begin{document}

%\maketitle
\thispagestyle{empty}
\pagestyle{empty}

%%%%%%%%%%%%%%%%%%%%%%%%%%%%%%%%%%%%%%%%%%%%%%%%%%%%%%%%%%%%%%%%%%%%%%%%%%%%%%%%

\twocolumn[{%
\renewcommand\twocolumn[1][]{#1}%
\maketitle
\begin{center}
    \centering
    \captionsetup{type=figure}
    \includegraphics[width=1.0\textwidth]{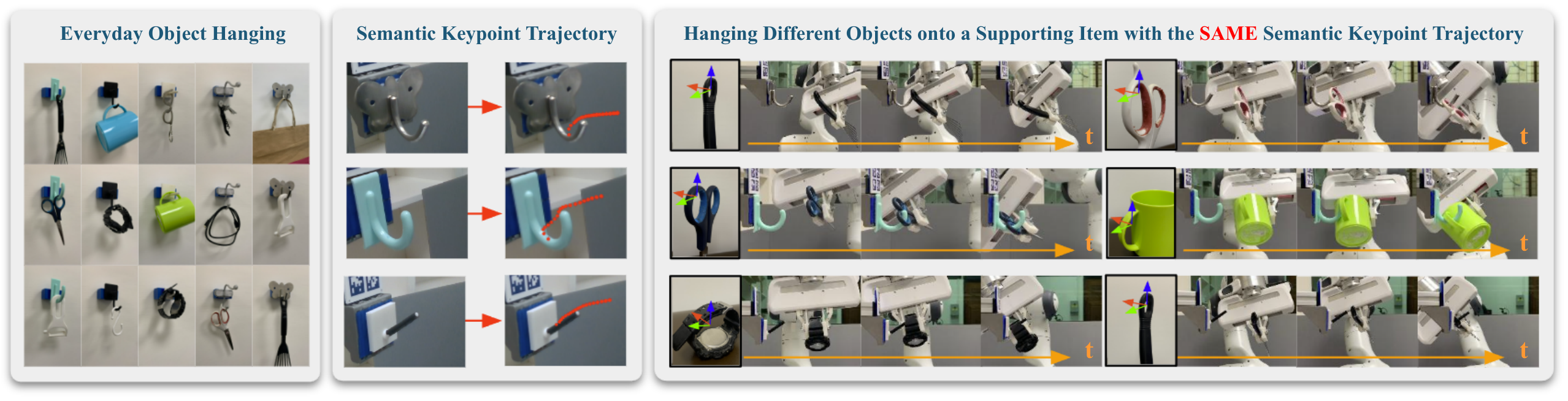}
    \captionof{figure}{
        The task of robot hanging a grasped object onto a supporting item involves a diverse range of grasped objects and supporting items with various shapes and geometric structures.
        %, making it challenging to determine the appropriate actions for successful hanging.
        %
        In this work, we introduce \textit{\textbf{S}emantic \textbf{K}eypoint \textbf{T}rajectory} (SKT), an actionable representation that specifies \textbf{where} and \textbf{how} to hang grasped objects onto a supporting item.
        Our experiments show that SKT is object-agnostic, enabling hanging various objects onto the same supporting item. 
        %to be applied across various objects.
        % while being \textcolor{red}{object-agnostic}.
    }
    \label{fig:overall}
\end{center}%
\vspace{-2mm}
}]

\begin{abstract}

We study the problem of hanging a wide range of grasped objects on diverse supporting items.
%
% Hanging objects is an omnipresent task in numerous areas of our daily life.
%
Hanging objects is a ubiquitous task that is encountered in numerous aspects of our everyday lives.
However, both the objects and supporting items can exhibit substantial variations in their shapes and structures, bringing two challenging issues: (1) determining the task-relevant geometric structures across different objects and supporting items, and (2) identifying a robust action sequence to accommodate the shape variations of supporting items.
%
% (1) where are the task-relevant geometry structures across different objects and supporting items (2) how to find a robust action sequence for successful object hanging that adapts to the shape variations of supporting items. 
%
% To this end, we propose \textit{Semantic Keypoint Trajectory}, an actionable representation that models the task-relevant geometric
% structure of a target supporting item and defines the appropriate action sequence for hanging objects.
% %
% Moreover, the representation is designed to be object-agnostic, significantly expanding its versatility and applicability to a wide range of everyday objects.
To this end, we propose
\textit{Semantic Keypoint Trajectory} (SKT), 
%
%we propose \textit{Semantic Keypoint Trajectory} (SKT), an actionable representation that models the hanging part of a target supporting item and defines suitable action sequences for hanging objects. 
%
an object-agnostic representation that is highly versatile and applicable to various everyday objects.
%
%This representation provides an intuitive and interpretable solution that specifies both where and how to hang while being object-agnostic, enabling the robot to efficiently complete the hanging task. 
% By leveraging the insight that a trajectory for hanging highly depends on the fine-grained geometry structure of a supporting item, we also introduce the \textit{Shape-Conditioned Trajectory Deformation Network (SCTDN)}
%To learn such representation, we also introduce the \textit{Shape-Conditioned Trajectory Deformation Network (SCTDN)}, a simple yet effective network that is able to generate feasible semantic keypoint trajectories for novel shapes by deforming an existing trajectory.
%
We also propose \textit{Shape-conditioned Trajectory Deformation Network} (SCTDN), a model that learns to generate SKT by deforming a template trajectory based on the task-relevant geometric structure features of the supporting items.
%the shape-conditioned point features of the targeted supporting item.
%
% The proposed SKT and SCTDN aim to tackle the challenges of determining the action sequences for hanging various objects onto diverse supporting items.
%other supporting items with similar geometric structures for the targeted supporting items.
% We propose a Shape-conditioned Trajectory Deformation Network (SCTDN) that learns to generate an action sequence by deforming a template trajectory derived from other supporting items with similar geometric structures for the targeted supporting items.     
%
We conduct extensive experiments and demonstrate substantial improvements in our framework over existing robot hanging methods in the success rate and inference time.
%
%\textcolor{blue}{
% Furthermore, we demonstrate the substantial potential of applying our simulator-trained framework to real-world scenarios for hanging grasped objects on diverse supporting items.
%We demonstrate the versatility and applicability of translating our simulator-trained framework into real-world scenarios for hanging grasped objects on diverse supporting items
%
Finally, our simulation-trained framework shows promising hanging results in the real world.
For videos and supplementary materials, please visit our project webpage: \href{https://hcis-lab.github.io/SKT-Hang/}{https://hcis-lab.github.io/SKT-Hang/}.

\end{abstract}

%%%%%%%%%%%%%%%%%%%%%%%%%%%%%%%%%%%%%%%%%%%%%%%%%%%%%%%%%%%%%%%%%%%%%%%%%%%%%%%%

\section{Introduction}
\label{sec:introduction}
%\textcolor{red}{motivation and the importance of the hanging task and challenges}

%
% Hanging objects is a common and everyday task in our lives.
Hanging is a common routine in various aspects of our daily lives, taking place in situations like households, logistics, and stationery stores.
%
% For instance, in household settings like kitchens, it is common to hang utensils, mugs, and scissors on wall hooks to save space and maintain organization.
% %
% Additionally, commercial venues such as shopping malls and stationery stores utilize hooks to display items, optimizing the use of available space.
%
% Equipping robots with the skill to hang objects in unconstrained settings can significantly impact these domains due to the tremendous potential to reduce human costs and enhance efficiency.
Equipping robots with the skill to hang objects in unconstrained settings can significantly alleviate labor shortages in these domains.
%
% In this paper, we study the problem of robotic hanging of a wide range of objects on diverse supporting items such as hooks and racks.
In this paper, we focus on how to enable robots to hang a wide range of objects in hand onto arbitrary supporting items, such as hooks or racks.
This is challenging since the robot needs to reason about the hanging structure and determine the corresponding actions while being robust to the diversity in the hanged objects and supporting items.

Existing algorithms~\cite{manuelli2019kpam, you2021omnihang, simeonov2022neural, pan2023tax} apply category-level representations, such as semantic keypoint or dense correspondence, to identify the hanging parts such as mug handles or contact points on supporting items. 
% %
These category-level representations are subsequently leveraged to define the desired target poses, demonstrating their utility in the task of hanging mugs on racks~\cite{manuelli2019kpam, simeonov2022neural, pan2023tax}.
%
%~\cite{manuelli2019kpam, simeonov2022neural, pan2023tax} employed these representations to identify elements like mug handles or contact points on supporting items for defining the desired target poses, showcasing their effectiveness in the task of hanging mugs on racks.
%
% Recent studies have explored representations such as semantic keypoints and dense correspondence to identify task-relevant object parts and establish the desired target configuration for manipulation tasks, including tool use~\cite{qin2020keto, gao2021kpam2}, grasping~\cite{florence2018dense}, object placement~\cite{manuelli2019kpam, simeonov2022neural, pan2023tax, xue2023useek}, and peg-hole insertion~\cite{gao2021kpam2}.
%
% In the context of hanging tasks,~\cite{manuelli2019kpam, simeonov2022neural, pan2023tax} utilized these representations to identify the hanging parts like mug handles and contact points on supporting items and identify the desired target poses for manipulation, showcasing their effectiveness in the task of hanging mugs on racks.
%
However, mere knowledge of hanging poses is insufficient when dealing with diverse objects and supporting items that require fine-grained intermediate actions (see Fig.~\ref{fig:overall}).

% \textcolor{blue}{It is crucial to determine a sequence of actions to hang objects on diverse supporting items.}https://www.overleaf.com/project/64fc9ecda867b815e0ca68e7
To determine action sequences for hanging diverse objects and supporting items, OmniHang~\cite{you2021omnihang} proposes a two-stage solution.
%
% Omnihang~\cite{you2021omnihang} tackles this challenge with their two-stage solution.
%
First, they detect contact points between objects and supporting items to identify suitable hanging poses for objects.
% The work most related to ours is OmniHang~\cite{you2021omnihang}, which employs contact points between objects and supporting items to identify suitable hanging poses for objects. 
% %
Second, they employ a sampling-based motion planner to generate collision-free trajectories for hanging.
% the objects
However, sampling-based motion planners require exhaustive search, resulting in significant computational burdens to determine a feasible trajectory for each object-supporting item pair. 
%
% Moreover, collision predicting under partial observability presents a significant challenge, resulting in trajectories that struggle to achieve collision-free outcomes.
Furthermore, predicting collisions under partial observability poses a significant challenge in generating feasible hanging trajectories.

% While existing approaches have shown promise in identifying the hanging pose across various objects and supporting items, they still face difficulties in identifying the suitable intermediate actions required for successful hanging.

% While keypoints provide insight into task-related functional parts across various objects and define the desired target configurations for task completion, they do not effectively guide robots in generating a sequence of feasible actions for executing the hanging task with a variety of target supporting items.

%\textcolor{red}{our proposed representation for dealing with the challenges and limitations of current works}

In this work, we present a new representation called \textit{\textbf{S}emantic \textbf{K}eypoint \textbf{T}rajectory} (SKT) and a novel framework for generating SKT to address the challenges faced in the existing methods.
SKT is an actionable representation that simultaneously models the hanging part of the supporting item and the hanging movements of the objects' keypoints. 
% To this end, we propose \textit{Semantic Keypoint Trajectory}, an actionable representation that simultaneously models the hanging part of the supporting item, and defines the appropriate action sequence for hanging objects.
%
Fig.~\ref{fig:overall} illustrates the concept of SKT.
%
%The inspiration 
%behind this idea 
%stems from the observation 
% We observe significant similarities in the movements of objects' keypoints when hanging onto the same supporting item, leading to the proposal of the SKT representation.
The design of SKT is inspired by the observation that keypoints can be defined across objects such that they have similar movements when hanging onto the same supporting item.
In our experiments, we validate that SKT has the capacity to facilitate the hanging of a wide range of objects onto the same supporting item. 
%universally applied across different objects for hanging.
%
It is worth noting that SKT can effectively eliminate the need for replanning for every object-supporting item pair, saving the computational resources for other prominent tasks.
%\textcolor{blue}{Note that the SKT is agnostic to objects, eliminating the need for any path re-planning when hanging a new object on the same supporting item.}
%
% \textcolor{red}{Taking advantage of this characteristic, we can address the challenge of efficiently identifying the appropriate actions for hanging objects on a variety of supporting items by generating the SKT for each supporting item.}
% Taking advantage of this object-agnostic characteristic of SKT, we propose a \textit{\textbf{S}hape-\textbf{C}onditioned \textbf{T}rajectory \textbf{D}eformation \textbf{N}etwork} (SCTDN) to obtain the SKT for each supporting item.
%
% This object-agnostic characteristic significantly enhances its versatility and applicability to a wide range of everyday objects.

% We propose a \textit{\textbf{S}hape-\textbf{C}onditioned \textbf{T}rajectory \textbf{D}eformation \textbf{N}etwork} (SCTDN) to obtain the SKT for each supporting item.
%
% Taking advantage of the object-agnostic characteristic of SKT and the proposed learning framework, we aim to tackle the challenge of effectively determining the appropriate actions for hanging various objects on diverse supporting items.
%
% Building on the well-established advancements in object keypoint prediction, our proposed representation and the learning framework aims to tackle the challenge of effectively determining the appropriate actions for object hanging tasks.

We introduce a novel learning framework for SKT generation: \textit{\textbf{S}hape-\textbf{C}onditioned \textbf{T}rajectory \textbf{D}eformation \textbf{N}etwork} (SCTDN).
%
% SCTDN takes the partial point cloud of a supporting item as input and generates the corresponding SKT by retrieving a template SKT from a trajectory database. 
% %
% It then refines this template SKT into the final SKT based on the point features corresponding to the hanging part of the input point cloud.
SCTDN takes the partial point cloud of a supporting item as input and generates the corresponding SKT by deforming a retrieved template SKT based on the task-relevant geometric structure features. 
We consider the following two critical design choices. 
%stems from two critical observations. 
%
%First, our empirical evidence suggests that deforming a trajectory is more effective than predicting one from scratch.
%
First, we suggest deforming a template trajectory rather than predicting it from scratch, based on the observation that similar SKTs correspond to similar hanging parts of supporting items.
We propose an unsupervised clustering algorithm to construct a template trajectory database. 
%Building on this insight, we adopt unsupervised categorization to group supporting items based on their SKTs, thus establishing a template trajectory database for template SKT retrievals.
%
% First, we suggest deforming a trajectory rather than predicting SKT from scratch because \textcolor{red}{Could you please add some insights?}.
%
% The second observation is that similar SKTs correspond to similar hanging parts of the supporting items. 
% %
% For example, when using curved hooks, the SKTs tend to follow curved paths, while straight hooks typically produce straight SKTs. 
% %
% Drawing inspiration from this observation, we implement unsupervised categorization to group supporting items based on their SKTs, creating a template trajectory database for template SKT retrievals.
%
Second, we extract features of task-relevant geometric structures when fine-tuning SKTs.
This approach is inspired by the trajectory generation with affordance guidance introduced in~\cite{wu2021vat}.

We demonstrate the effectiveness of SCTDN with a promising hanging success rate for a wide range of objects and supporting items compared to state-of-the-art algorithms~\cite{gao2021kpam2,you2021omnihang,wu2021vat}. 
We also show efficiently generated action sequences compared to~\cite{you2021omnihang}. 
Moreover, given a supporting item, the SKTs predicted by SCTDN can enable hanging various objects compared to a strong baseline~\cite{wu2021vat}.  
%We conduct extensive experiments to evaluate the hanging success rate and the inference time of the existing methods and our proposed method in the PyBullet simulator~\cite{coumans:2021}.
% 
Specifically, we apply all baselines to hang 50 different grasped objects onto 60 unseen supporting items, resulting in a total of 3,000 hanging trials.
All methods must determine the corresponding hanging actions for diverse pairs of objects and supporting items under partial observability, making our evaluation setting challenging.
%
% The results demonstrate that our proposed method achieves a significant improvement \textcolor{red}{in both success rate and inference time}, surpassing existing state-of-the-art algorithms.
% %
We perform thorough ablative studies to justify our design choices, i.e., the selection of deformation over prediction from scratch and the importance of integrating task-relevant geometric structures.
% %
Finally, we show promising results of hanging grasped objects onto various supporting items in the real world.
%More importantly, our simulator-trained framework shows promising results for translation into real-world scenarios in the context of hanging grasped objects onto various supporting items.
% Last but not least, given a generated semantic keypoint trajectory for a supporting item, we demonstrate the trajectory enables hanging diverse objects with a promising success rate.
%

The contributions of this paper are as follows:
\begin{itemize}
\item We present \textit{Semantic Keypoint Trajectory}, a novel actionable and object-agnostic representation that simultaneously models the hanging structure of the supporting item and the movements of the objects' keypoints.
% for identifying suitable action sequences of various objects for the hanging task. 

\item We introduce \textit{Shape-conditioned Trajectory Deformation Network (SCTDN)}, a novel framework for generating semantic keypoint trajectories through reasoning over the geometry of the hanging component and performing trajectory deformation.
%
%effectively generate semantic keypoint trajectory via explicit modeling of the geometry of the hanging part and trajectory deformation.  

\item We perform comprehensive experiments and demonstrate significant improvements in both the success rate and inference time, surpassing state-of-the-art object hanging algorithms. 

%\item We demonstrate the versatility and the applicability of translating our simulator-trained framework in the real world.
\item We present compelling results of hanging grasped objects onto various supporting items in the real world.
% in the context of hanging grasped objects onto various supporting items.

\end{itemize}

\section{Related Work}
\label{sec:related_work}

\subsection{Keypoint in Robot Manipulation}

 % Recent studies have explored representations such as semantic keypoints and dense correspondence to identify task-relevant object parts and establish the desired target configuration for manipulation tasks, including tool use~\cite{qin2020keto, gao2021kpam2}, grasping~\cite{florence2018dense}, object placement~\cite{manuelli2019kpam, simeonov2022neural, pan2023tax, xue2023useek}, and peg-hole insertion~\cite{gao2021kpam2}.

Keypoint has gained significant popularity in the realm of generalized robot manipulation~\cite{florence2018dense, manuelli2019kpam, qin2020keto, gao2021kpam2, vecerik2021s3k, pan2023tax, xue2023useek}. 
% for recognizing the task-relevant geometric structures across various objects.
%
These approaches have yielded promising results in manipulation tasks such as grasping~\cite{florence2018dense}, tool use~\cite{qin2020keto, gao2021kpam2}, object placement~\cite{manuelli2019kpam, simeonov2022neural}, mug hanging~\cite{manuelli2019kpam, pan2023tax}, and peg-hole insertion~\cite{gao2021kpam2}.
%
% However, modeling task-relevant geometric structures alone falls short in the context of hanging, as it remains unclear how to determine the corresponding action sequences for objects onto a diverse set of supporting items.
Recent studies have also delved into integrating keypoints with reinforcement learning or imitation learning to determine robot actions for long-horizon tasks such as object pushing~\cite{florence2019self, manuelli2020keypoints}, pick and place~\cite{florence2019self}, shoe flipping~\cite{florence2019self}, and cable insertion~\cite{vecerik2021s3k}.
Inspired by the success of keypoint in modeling task-relevant geometric structures across different objects, we introduce the Semantic Keypoint Trajectory as a novel actionable representation for the hanging task. 
We demonstrate its effectiveness in offering sequential guidance for the hanging of various objects onto a variety of supporting items.

\subsection{Actionable Representations for Affordance}

In recent years, the research community has been exploring actionable representations, such as the combination of affordance and trajectory~\cite{mo2021where2act, wu2021vat, xu2022universal, liu2022joint, bahl2023affordances}, optical flow~\cite{weng2022fabricflownet}, and scene flow~\cite{eisner2022flowbot3d, seita2023toolflownet}, for various downstream applications, including articulated object manipulation~\cite{mo2021where2act, wu2021vat, xu2022universal, bahl2023affordances, eisner2022flowbot3d}, hand motion prediction~\cite{liu2022joint}, fabric manipulation~\cite{weng2022fabricflownet}, and tool use~\cite{seita2023toolflownet}.
The goal is to bridge the perception-action gap, enhancing the ability to transfer manipulation skills to unseen configurations. 
A crucial difference between our proposed representation and existing works lies in our emphasis on the trajectory of the task-relevant geometric structures of an object, rather than the hand pose or gripper pose of a robotic arm.

Among these works, our proposed framework exhibits similarities to the \textit{object-centric actionable visual priors} proposed in VAT-Mart~\cite{wu2021vat}. 
VAT-Mart jointly learns the interaction point given a partial point cloud and the corresponding trajectory for articulated object manipulation.
%in 3D space 
In our work, we introduce a trajectory deformation-based approach, instead of a trajectory prediction framework.
Empirically, we demonstrate this strategy leads to more effective and efficient trajectory generation for diverse hanging tasks.

\section{Semantic Keypoint Trajectory and Dataset}
\label{chapter:template_trajectory_database}

\subsection{Semantic Keypoint Trajectory (SKT)} 
\label{section:semantic_keypoint_trajectory}

% We propose a new representation called \textit{Semantic Keypoint Trajectory}.
%for the task of everyday object hanging.
%
A semantic keypoint trajectory is defined as $\mathcal{T} = \{\xi_0, \xi_1,\ldots,\xi_{T - 1}\}$, where $\mathcal{T}$ is a sequence of 6D poses defined in a supporting item’s coordinate frame, $\xi_i \in \text{SE}(3)$ is the $i$'th waypoint, and $T$ is the trajectory length.
It is worth noting that, $\xi_0$ is the end waypoint attached to the supporting item.
%
% \textcolor{blue}{The semantic keypoint trajectory of a supporting item can be viewed as a "canonical trajectory" for hanging different objects by following the trajectory.}

\begin{figure}[t!]
\centering
    \includegraphics[width=1.0\columnwidth,clip]{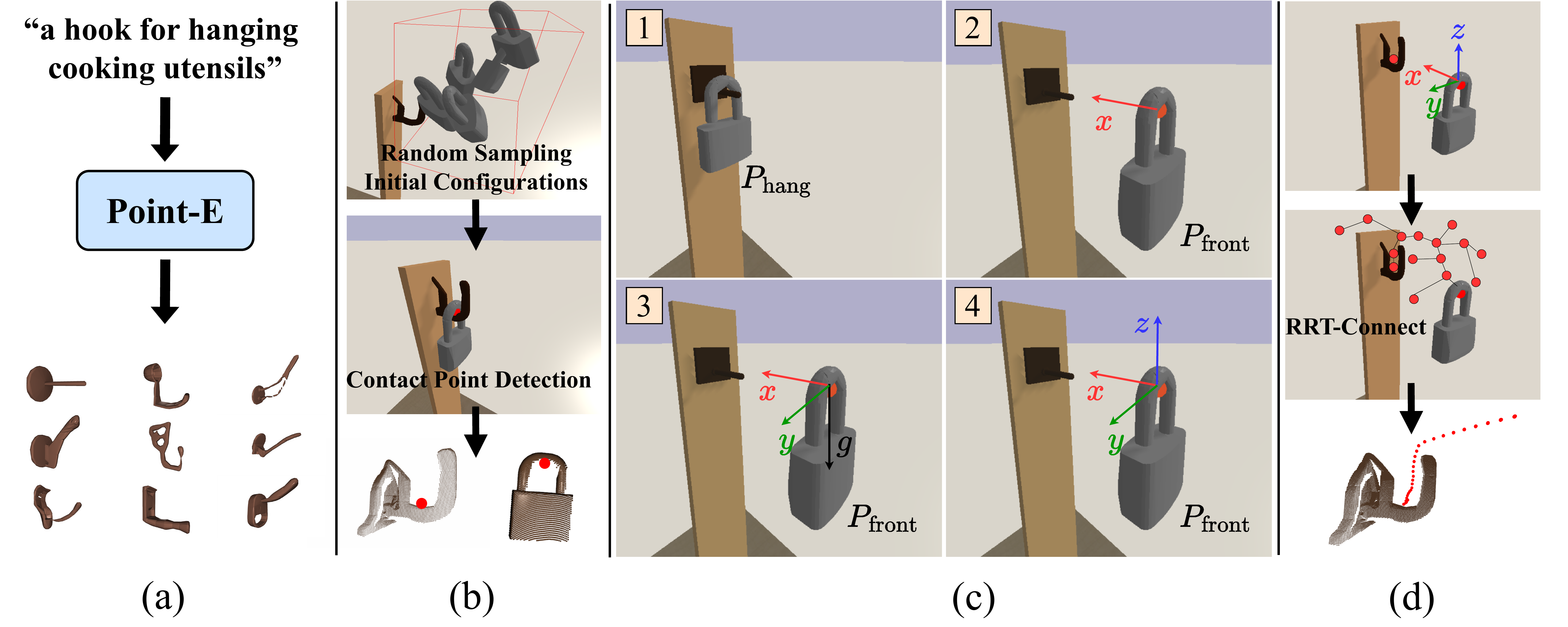}
    \caption{
        The overall data collection pipeline. (a) Shape generation via text-to-3D framework~\cite{nichol2022point}. (b) Contact point collection via forward simulation. (c)  Rotation extraction via the object's forward direction and the gravity direction. (d) Semantic keypoint trajectory generation via RRT-Connect~\cite{rrt-connect}. 
        % (e) Affordance map generation using the contact point and the semantic keypoint trajectory. (f) Unsupervised categorization to group the supporting items by the corresponding trajectories.
    }
    \label{fig:dataset}
\vspace{-7mm}
\end{figure}

\noindent\textbf{Rotation Definition.} We systematically define the rotation of a semantic keypoint $\xi_i \in \text{SE}(3)$ as follows (see Fig.~\ref{fig:dataset} (c)):
We define the forward direction of an object hung onto the supporting item as the x-axis of a semantic keypoint $\xi_i$.
The y-axis of a semantic keypoint $\xi_i$ is the cross product between the x-axis of a semantic keypoint and the unit vector of gravity.
The z-axis is subsequently defined as the cross product of the unit vectors the x-axis and y-axis. 
There are two notable observations that elucidate the basis for defining rotations in this specific manner.
%
%First, in the hanging task, we observed that the direction of object movement along the trajectory is usually perpendicular to the plane formed by the hanging parts on the object such as the "hanging ring" at the tail of a tool or the mug handle.
%
First, in the hanging task, we observed that the direction of object movement is usually perpendicular to the plane formed by the hanging part on an object, such as the "hanging ring" at the tail of a tool or the mug handle. 
%
% Second, the design choice streamlines the automatic collection of rotations during the data collection process, \textcolor{red}{please refer to~\ref{subsec:data_collection} for more details.}
%
% Third, this design choice offers a significant advantage because it allows for the simplification of the trajectory to $T \times 3$ dimensions during training, retaining only the positional component, thereby eliminating the complexity of regressing a sequence of $SO(3)$ rotations.
% Second, this design choice offers a significant advantage because it simplifies the trajectory representation,
Second, this design choice offers a significant advantage because it allows for the simplification of the trajectory representation during training, 
% to $3$ dimensions during training, 
retaining only the positional component that eliminates the complexity of regressing a sequence of \text{SO}(3) rotations.
During inference, we use adjacent 3D waypoints to identify the moving direction (i.e., the x-axis).
%gravity direction to generate the corresponding $\text{SE(3)$ waypoints.
%
Empirically, we demonstrate the design choice can favorably hang a wide range of objects to diverse supporting items.

% To complete a hanging manipulation, the robot follows a semantic keypoint trajectory, starting from $\xi_{T-1}$ and progressing towards $\xi_0$ (in reverse order of $\mathcal{T}$).
\noindent\textbf{Task execution.} First, the relative transform between the robot's end-effector pose $T^{\text{ee}}$ and the grasped object's \text{SE}(3) keypoint $T^{\text{kpt}}$ is recorded as $T^{\text{ee}}_{\text{kpt}} = (T^{\text{kpt}})^{-1} T^{\text{ee}}$.
We assume $T^{\text{ee}}_{\text{kpt}}$ remains fixed after an object is stably grasped.
% %
This assumption is aligned with~\cite{gao2021kpam2}. 
To obtain $T^{\text{kpt}}$ for each object, one can use off-the-shelf methods such as~\cite{gao2021kpam2,takeuchi2021automatic}.
%Gao and Tedrake~\cite{gao2021kpam2} and Takeuchi et al.~\cite{takeuchi2021automatic}.
%, and the relative transform $T^{\text{ee}}_{\text{kpt}} = (T^{\text{kpt}})^{-1} T^{\text{ee}}$ is recorded, where $T^{\text{kpt}}$ is the oriented keypoint of the grasped object, and $T^{\text{ee}}$ is the end-effector pose of the robot arm.
% %
% One can use off-the-shelf methods such as Wei Gao, Russ Tedrake~\cite{gao2021kpam2} and Takeuchi et al.~\cite{takeuchi2021automatic} to obtain the oriented object keypoints.
% %
Then, the robot follows a semantic keypoint trajectory, starting from $\xi_{T-1} \times T^{\text{ee}}_{\text{kpt}}$ and progressing towards $\xi_{0} \times T^{\text{ee}}_{\text{kpt}}$ (in reverse order of $\mathcal{T}$) to complete a hanging manipulation.
% Then, the hanging process is completed as the robot reaches each gripper pose $\xi_i \times T^{\text{ee}_{\text{kpt}}}$ corresponding to the waypoints of the semantic keypoint trajectory from $\xi_{T-1}$ to $\xi_0$.

\begin{figure*}[t!]
\centering
\hspace{10mm}
    \includegraphics[width=0.9\textwidth]{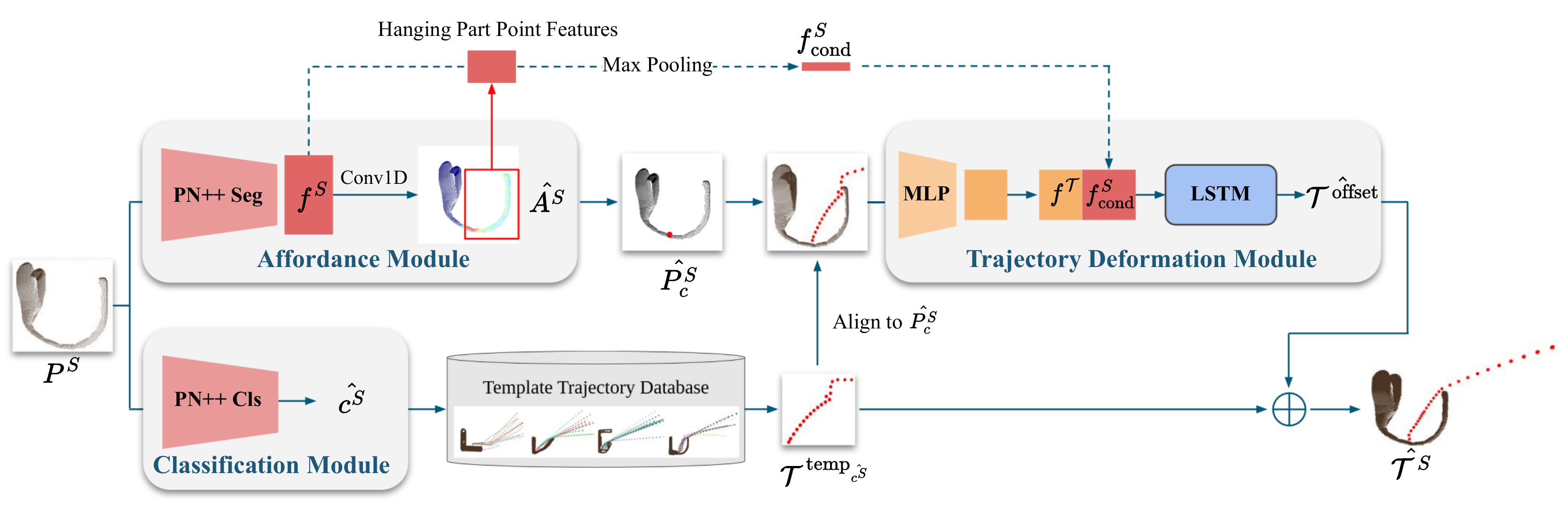}
    \caption{The architecture of the proposed \textbf{S}hape-\textbf{C}onditioned \textbf{T}rajectory \textbf{D}eformation \textbf{N}etwork (SCTDN). }
    \label{fig:arch}
\vspace{-5mm}
\end{figure*}

\subsection{Data Collection Pipeline}
\label{subsec:data_collection}

To learn to predict SKT, we propose an automated data collection pipeline within a simulation environment. 
We make this choice due to the challenges and costs associated with collecting a substantial number of supporting items, along with their corresponding semantic keypoints and SKTs in the real world. The data collection pipeline is illustrated in Fig.~\ref{fig:dataset}.

First, we apply Point-E~\cite{nichol2022point}, a text-to-3D framework, to collect a diverse set of supporting items (see Fig.~\ref{fig:dataset}(a)).
To obtain the semantic keypoints, we automatically collect the contact points of the supporting items and the hanging objects as semantic keypoints via forward simulation (see Fig.~\ref{fig:dataset}(b)), as demonstrated in~\cite{you2021omnihang,takeuchi2021automatic}. 
% Similar to the existing works~\cite{you2021omnihang, takeuchi2021automatic}, our work also designates contact points as semantic keypoints.
%
% A contact point is a point where an object contacts a supporting item at a stable hanging pose, which can be obtained automatically via forward simulation, as demonstrated in~\cite{you2021omnihang,takeuchi2021automatic}. This process is illustrated in Fig.~\ref{fig:dataset} (a)."
% 

Fig.~\ref{fig:dataset}(c) demonstrates how we determine an object's semantic keypoint rotation. 
Using a horizontal supporting item during the contact point collection process, we first obtain the position $P_{\text{hang}}$ of the object hanging pose. 
We then move the object in front of the support item at the position $P_{\text{front}}$ and calculate the x-axis (i.e., forward direction) of the keypoint's rotation matrix as the unit vector of ($P_{\text{hang}} - P_{\text{front}}$). 
Next, we compute the y-axis of the keypoint's rotation matrix as the cross-product of the x-axis and the gravity unit vector.
Finally, we derive the z-axis as the cross-product of the x-axis and y-axis.
Given the availability of complete shapes of supporting items in simulation, we collect SKTs for a supporting item by employing the RRT-Connect path planner~\cite{rrt-connect} with the collision detection API in the Pybullet simulator~\cite{coumans:2021}.
%
% \textcolor{red}{Note that, the rotation of the corresponding semantic keypoint of the hanging objects is obtained in this process.}
% we assign the x-axis as $\frac{\vec{x}_{\text{avg}}}{|\vec{x_{\text{avg}}}|}$, where $\vec{x}_{\text{avg}} $ is the average of the relative forward vector in a semantic keypoint trajectory, denoted as $\vec{x}_{\text{avg}} = \frac{\sum_{i=1}^{T-1} \xi_i - \xi_{i-1}}{T-1}$.
%We take the average of the differences between adjacent waypoints as the x-axis
It is noteworthy that, we utilize a single object as the reference object to collect the SKTs for all the supporting items, as illustrated in Fig.~\ref{fig:dataset} (d). 
We select this object as the pivot object because its shape is relatively simple, enabling the RRT-Connect path planner to generate paths more efficiently. 
% We selected this object as the pivot object because its hanging part is small to ensure the reusability of the collected trajectories by other objects. 
% %
% Furthermore, its shape is relatively simple, enabling the RRT-Connect path planner to generate paths more efficiently. 

In summary, our dataset comprises 472 supporting items and 50 objects with semantic keypoints. 
We collected 50 semantic keypoint trajectories for each supporting item and captured 42 point clouds from various camera viewpoints. 
Our training set consists of 352 supporting items, while we reserved 60 supporting items for validation and another 60 for testing.

\section{Problem Formulation}
\label{sec:problem}

We consider the task of a robot hanging a grasped object onto a supporting item.
In this task, our primary objective is to predict the semantic keypoint trajectory, given the partial point cloud of a supporting item $\mathcal{P}^{S} \in \mathbb{R} ^ {N \times 3}$ as input. The parameter $N$ is the number of points of the supporting item. This trajectory then serves as a guide for objects to execute a sequence of actions for hanging.
%
%The input is the partial point cloud of a supporting item $\mathcal{P}^{S} \in \mathbb{R} ^ {N \times 3}$, captured from a single depth camera.
%
The output is a position-only semantic keypoint trajectory $\hat{\mathcal{T}^S} \in \mathbb{R}^{T \times 3}$ for the supporting item, where $T$ is the length of the trajectory.
During inference, the position-only trajectory is augmented to a SE(3) trajectory based on the method described in~\ref{section:semantic_keypoint_trajectory}.
%, which enables hanging the object to the supporting item by following the trajectory.}

% \chapter{Section Ordering}
% \label{chapter:secorder}

% Section ordering in \textit{thesis.cls} is:
% \begin{itemize}
% \item Chapter (shown in \textbf{Table of Contents})
% \item Section (shown in \textbf{Table of Contents})
% \item Subsection (shown in \textbf{Table of Contents})
% \item Paragraph
% \item Subparagraph
% \end{itemize}
% DONOT use \textbackslash subsubsection, it is not supported in \textit{thesis.cls}.
% It is replaced by \textbackslash paragraph.

% \section{Section}
% \label{sec:secorder}
% \lipsum[1-2]  % dumy text generator

% \subsection{Subsection}
% \label{subsec:secorder}
% \lipsum[3-4]  % dumy text generator

% \paragraph{Paragraph}
% \lipsum[5-6]  % dumy text generator

% \subparagraph{Subparagraph}
% \lipsum[7]  % dumy text generator

% \section{Section}
% \label{sec:secorder}
% \lipsum[8]  % dumy text generator

\section{Methodology}
\label{sec:method}

\subsection{Overview of SCTDN} 
\label{subsec:method}

We propose a \textbf{S}hape-\textbf{C}onditioned \textbf{T}rajectory \textbf{D}eformation \textbf{N}etwork (SCTDN), a novel learning framework for the hanging task. 
It leverages the strategy of deforming a template trajectory based on shape-conditioned features to generate the target semantic keypoint trajectory (SKT) for the partial point cloud of the input supporting item.

Fig.~\ref{fig:arch} offers an overview of the SCTDN architecture, which encompasses three core modules:

\noindent\textbf{Classification Module.} This module predicts the shape category $\hat{c^S}$ of the input point cloud $\mathcal{P}^S$.
It enables the retrieval of the appropriate template SKT $\mathcal{T}^{\text{temp}_{\hat{c^S}}}$ from the template trajectory database.
% 
% To select the template trajectory, we retrieve a semantic keypoint trajectory from a template trajectory database that corresponds to the same shape group as the input point cloud through a point cloud classification module.
%
This design choice is motivated by the observation that similar SKTs correspond to similar hanging parts of the supporting items. 
For example, when using curved hooks, the SKTs tend to follow curved paths, while straight hooks typically produce straight SKTs.
Inspired by this observation, we implement unsupervised categorization to group supporting items based on their SKTs and create a template trajectory database. 
Please refer to section~\ref{template_trajectory_database} for more details.

\noindent\textbf{Affordance Module.} This module predicts the affordance map $\hat{A^S} \in \mathbb{R}^{N \times 1}$ for the input point cloud $\mathcal{P}^S$. 
An affordance map is a heatmap that specifies the probability of each point belonging to the hanging part of a supporting item's point cloud.
The point with the maximal probability is the contact point $\hat{\mathcal{P}^S_c} \in \mathbb{R}^3$, which enables the alignment of the first waypoint of the template trajectory.
%
% The shape-conditioned feature used to guide the trajectory deformation comprises the point features corresponding to the hanging part of the input point cloud, which is obtained from a point cloud segmentation-based affordance prediction module.

\noindent\textbf{Trajectory Deformation Module.} This module predicts the waypoint offsets $\hat{\mathcal{T}^{\text{offset}}} \in \mathbb{R}^{T \times 3}$ to refine the template trajectory $\mathcal{T}^{\text{temp}{\hat{c^S}}}$ based on the shape-conditioned feature $f^S_{\text{cond}}$.
% corresponding to the hanging part of the input point cloud $\mathcal{P}^S$.
%
We obtain the shape-conditioned feature $f^S_{\text{cond}}$ by max-pooling the point features corresponding to the hanging part of the input point cloud from the Affordance Module.
The reason for including the shape-conditioned feature in trajectory deformation is based on the observation that there is a significant correlation between the SKT and the hanging part of a supporting item.

\noindent\textbf{Inference.} Given the partial point cloud $\mathcal{P}^{S}$ as input, the SCTDN first retrieves a template trajectory $\mathcal{T}^{\text{temp}_{\hat{c^S}}} = \{\xi^{\text{temp}_{\hat{c^S}}}_0, \xi^{\text{temp}_{\hat{c^S}}}_1,\ldots,\xi^{\text{temp}_{\hat{c^S}}}_{T-1}\} \in \mathbb{R}^{T \times 3}$ associated with the shape category $\hat{c^S}$ predicted by the classification module from the template trajectory database. 
%
% Then, the model predicts the contact point $\hat{\mathcal{P}^S_c}$ by choosing the index of the point with the highest affordance score, predicted by the affordance module. 
Then, the model determines the contact point $\hat{\mathcal{P}^S_c}$ by selecting the index of the point with the highest probability in $\hat{A^S}$, predicted by the Affordance Module.
We align the first waypoint of the template trajectory $\mathcal{T}^{\text{temp}_{\hat{c^S}}}$ to the contact point $\hat{\mathcal{P}^S_c}$ by adding $\xi^{\text{temp}_{\hat{c^S}}}_{0} - \hat{\mathcal{P}^S_c}$ to each waypont.
Subsequently, the trajectory deformation module takes the aligned $\mathcal{T}^{\text{temp}_{\hat{c^S}}}$ and a shape-conditioned feature $f^S_{\text{cond}}$ obtained from the affordance module as input and generate the waypoint offsets $\hat{\mathcal{T}^{\text{offset}}} = \{\hat{\xi^{\text{offset}}_0}, \hat{\xi^{\text{offset}}_1},\ldots,\hat{\xi^{\text{offset}}_{T-1}}\}\in \mathbb{R}^{T \times 3}$.
The final trajectories is defined as $\hat{\mathcal{T}^{S}} = \big\{ \{ \xi^{\text{temp}_{\hat{c^S}}}_{i} + \hat{\xi^{\text{offset}}_{i}}, \} | i \in [0, 1, ..., T-1] \big\}$.

\begin{table*}[t!]
\centering
\scriptsize
\caption{The success rate evaluation of each approach. \dag~denotes our re-implementation of the baseline methods.}
\resizebox{0.8 \textwidth}{!}{
        \begin{tabular}
            {@{} l @{\;} @{\;} | c @{\;} | c @{\;} | c @{\;} c @{\;} c @{\;} c @{\;}}
            \toprule
            & 
            \multicolumn{1}{c}{Inference Time (sec)}  & 
            \multicolumn{1}{c}{All (\%)}  & 
            \multicolumn{1}{c}{Easy (\%)}  & 
            \multicolumn{1}{c}{Normal (\%)}  & 
            \multicolumn{1}{c}{Hard (\%)}  & 
            \multicolumn{1}{c}{Very Hard (\%)} 
            \\
            \midrule
             K-PAM 2.0\dag~\cite{gao2021kpam2} & - & 40.1 & 76.0 & 54.1 & 21.6 & 8.8
            \\
             OmniHang\dag~\cite{you2021omnihang} & 14.29 & 40.1 & 62.9 & 47.7 & 34.3 & 15.3
            \\
             \midrule
             VAT-Mart\dag~\cite{wu2021vat} & \textbf{0.007} & 55.2 & 84.5 & 82.0 & 42.0 & 12.3
            \\
             Modified VAT-Mart \dag & 0.016 & 72.1 & 78.8 & 89.6 & 57.7 & 62.4
            \\
             \midrule
              SCTDN (Ours) & 0.009 & \textbf{83.7} & \textbf{85.5} & \textbf{91.7} & \textbf{79.7} & \textbf{77.7}
            \\
            \bottomrule
        \end{tabular}
}
\label{table:main}
\vspace{-5mm}
\end{table*}

\subsection{Template Trajectory Database and Affordance Maps} 

\begin{figure}[t!]
\centering
    \includegraphics[width=1.0\columnwidth,clip]{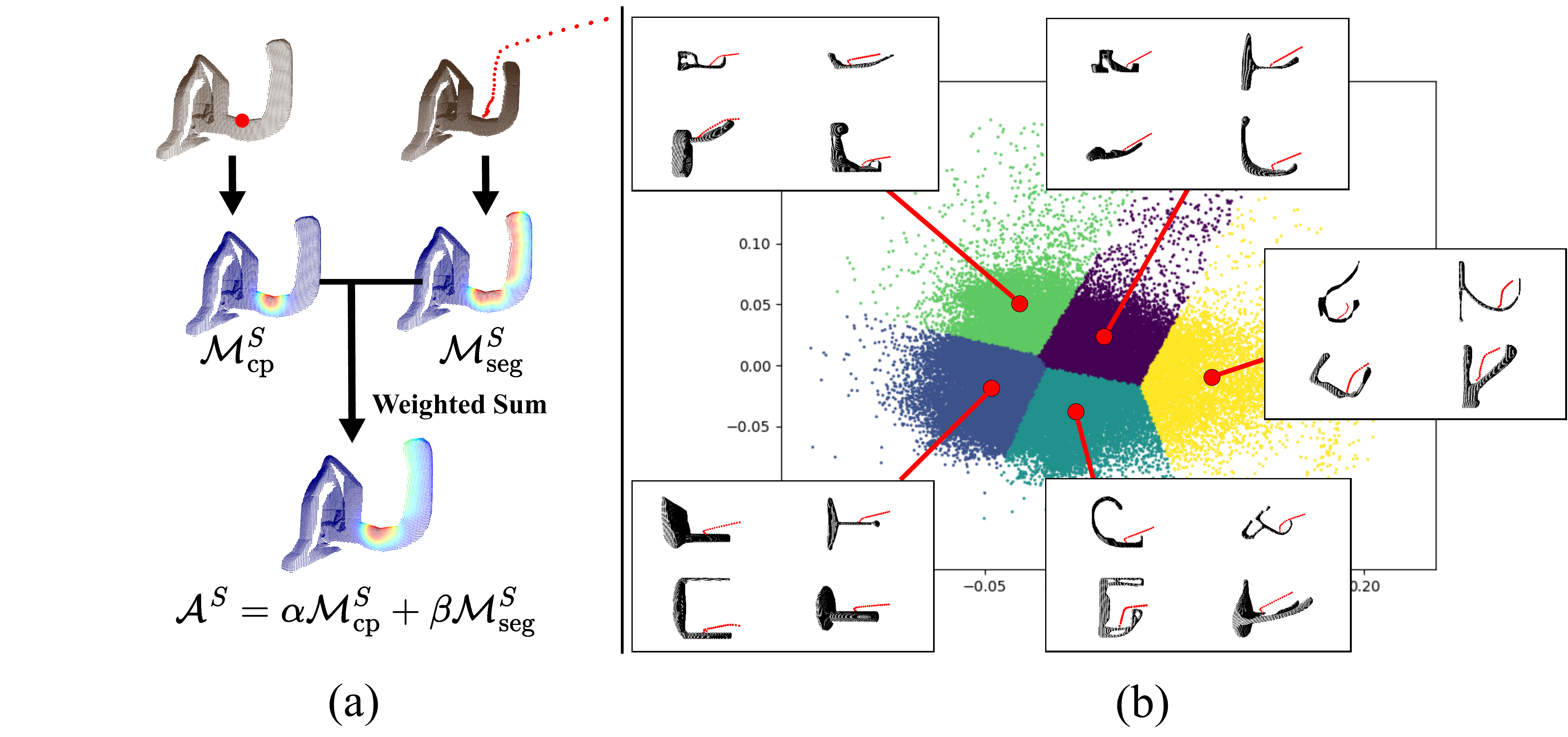}
    \caption{
         The data post-processing pipeline. (a) Affordance map generation pipeline.
         %using the contact point and the semantic keypoint trajectory. 
         (b) Unsupervised categorization to group the supporting items by the corresponding trajectories.
    }
    \label{fig:post_processing}
\vspace{-6mm}
\end{figure}

\noindent\textbf{Template Trajectory Database Creation.} First, we flatten each trajectory $\mathcal{T}$ in the training set and reduce its dimensionality to 2 using principal component analysis.
\label{template_trajectory_database}
Following that, we apply the K-means clustering algorithm to cluster the lower dimensional vectors into groups (see Fig.~\ref{fig:post_processing} (b)). 
Finally, the category label $c^S \in [0, K-1]$ is assigned to each supporting item based on the closest center.
With the above process, we build a template trajectory database for the development of the proposed framework.
This database contains the SKTs corresponding to the supporting item that is closest to each center. 
%
% The visualization of the clustering can be found in Fig.~\ref{fig:dataset} (d).

\noindent\textbf{Affordance Maps Generation.} To facilitate the training of the affordance prediction module, we devise an automatic post-processing pipeline for generating the ground truth affordance maps. 
%
% An affordance map is a heatmap that specifies the probability of each point belonging to the hanging part of a supporting item's point cloud. 
%
% The point with the maximal probability is the contact point, which enables the alignment of the first waypoint of the template trajectory.
%
Fig.~\ref{fig:post_processing} (a) illustrates the generation process.
Initially, we identify the points on the partial point cloud closest to the contact point and each waypoint within the SKT. 
By leveraging mixed Gaussian distributions, we generate two distinct probability maps for each point in a point cloud - the contact map $\mathcal{M}_{\text{cp}}^S$, and the segmentation map $\mathcal{M}_{\text{seg}}^S$.
The final affordance map is then rendered by employing a weighted sum of the two maps, denoted as $\mathcal{A}^S = \alpha \mathcal{M}_{\text{cp}}^S + \beta \mathcal{M}_{\text{seg}}^S$. 
Practically, we set $\alpha = 0.5$ and $\beta = 0.5$, and consider the variance $\sigma^2$ as a hyperparameter (in our case, it's set to 0.5 cm).

% By leveraging mixed Gaussian distributions, we generate two distinct probability maps for each point in a point cloud - the contact map $\mathcal{M}_c(\mathcal{P}^s_i)$, and the segmentation map $\mathcal{M}_s(\mathcal{P}^s_i)$:

% \begin{equation}
% \footnotesize
% \mathcal{M}_c(\mathcal{P}^s_i) = \frac{N(\mathcal{P}^s_i; \mathcal{P}^s_c, \sigma^2)}{\mathop{\arg\max}\limits_{j} N(\mathcal{P}^s_j; \mathcal{P}^s_c, \sigma^2)}, \forall \mathcal{P}^s_i \in \mathcal{P}^s
% \end{equation}

% \begin{equation}
% \footnotesize
% \mathcal{M}_s(\mathcal{P}^s_i) = \frac{\mathop{\arg\max}\limits_{\xi_t} N(\mathcal{P}^s_i; \mathcal{P}^s_{\xi_t},\sigma^2)}{\mathop{\arg\max}\limits_{j, \xi_t} N(\mathcal{P}^s_j; \mathcal{P}^s_{\xi_t}, \sigma^2) }, \forall \mathcal{P}^s_i \in \mathcal{P}^s
% \end{equation}

% Here, $\mathcal{P}^s_i$ represents each point on the supporting item's point cloud $\mathcal{P}^s$, $\mathcal{P}^s_c$ denotes the contact point on the point cloud, and $\mathcal{P}^s_{\xi_t}$ corresponds to the closest point to the $t$’th waypoint on the point cloud. 
% %
% The final affordance map is then rendered by employing a weighted sum of the two maps, denoted as $\mathcal{M}_a(\mathcal{P}^s_i) = \alpha \times \mathcal{M}_c(\mathcal{P}^s_i) + \beta \times \mathcal{M}_s(\mathcal{P}^s_i)$. 
% %
% Practically, we set $\alpha = 0.5$ and $\beta = 0.5$, and consider the variance $\sigma^2$ as a hyperparameter (in our case, it's set to 0.5 cm). 

\subsection{Implementation Details} 
% Figure~\ref{fig:arch} (a) offers an overview of the SCTDN architecture, which encompasses three core modules:

In our experiments, we configure the point number $N=1000$, and the category number $K=5$, and the trajectory length $T \in \{10, 20, 40\}$.
We identify the hanging part as the point with a probability greater than 0.1 in $\hat{A^S}$.

\noindent\textbf{Classification Module.} %This module predicts the shape category $\hat{c^S}$ of the input point cloud $\mathcal{P}^S$. 
%
% It enables the retrieval of the appropriate template semantic keypoint trajectory $\mathcal{T}^{\text{temp}_{\hat{c^S}}}$ from the template trajectory database.
% For training, 
We apply a \textit{classification} PointNet++~\cite{qi2017pointnet++} network to extract the global feature $f^g \in \mathbb{R} ^ {512}$, then input it to a two-layer MLP network $(512 \rightarrow 256 \rightarrow K)$ to predict the class $\hat{c^S}$. 
The loss used for this module is the standard cross-entropy loss, denoted as $L_c$.

\noindent\textbf{Affordance Module.} %This module predicts the affordance map $\hat{A^S} \in \mathbb{R}^{N \times 1}$ for the input point cloud $\mathcal{P}^S$. 
%
% The point with the maximal affordance score is the contact point of the input point cloud, denoted as $\hat{P^S_c}$, which enables the alignment of the first waypoint of the template trajectory $\mathcal{T}^{\text{temp}_{\hat{c^S}}}$ during inference.
%
% This map specifies the likelihood of points belonging to the hanging part or contact point.
% For training, 
We apply a \textit{segmentation} PointNet++~\cite{qi2017pointnet++} network to extract the per-point feature $f^S \in \mathbb{R}^{N \times 512}$ of the input point cloud. 
Subsequently, given the $f^S$, we apply a Conv1D layer along with the sigmoid function to predict the affordance map $\hat{A^S}$.
We utilize the standard L2 loss as the affordance loss, denoted as $L_a$.
% We formulate affordance prediction as a binary classification problem. Thus, we utilize the binary cross-entropy loss as the affordance loss, denoted as $L_a$.

\noindent\textbf{Trajectory Deformation Module.} %This module predicts waypoint offsets $\hat{\mathcal{T}}^{\text{offset}}$ to refine the template trajectory $\mathcal{T}^{\text{temp}{\hat{c^S}}}$ based on the shape-conditioned feature $f^S_{\text{cond}}$ corresponding to the hanging part of the input point cloud $\mathcal{P}^S$.
During the training stage, the inputs for this module consist of a template SKT $\mathcal{T}^{\text{temp}_{c^S}}$ correspond to the input shape category $c^S$ and a shape-conditioned feature $f^S_{\text{cond}}$ derived from the per-point features $f^S$.
%
% To obtain the shape-conditioned feature $f^S_{\text{cond}} \in \mathbb{R}^{N \times 512}$, we perform max-pooling on the point features associated with points having affordance scores exceeding a specified threshold (typically set to 0.1). These points represent the hanging part of the input point cloud.
%
First, we apply a three-layer MLP network $(T\times 3 \rightarrow 64 \rightarrow 32 \rightarrow 32)$ to encode each waypoint in $\mathcal{T}^{\text{temp}_{c^S}}$, resulting in waypoint features $f^{\mathcal{T}} \in \mathbb{R}^{T \times 32}$. 
Subsequently, we concatenate the shape-conditioned feature $f^S_{\text{cond}}$ with each waypoint feature in $f^{\mathcal{T}}$. 
These combined features serve as inputs to two LSTM~\cite{hochreiter1997long} blocks, ultimately yielding predicted waypoint offsets $\hat{\mathcal{T}}^{\text{offset}}$. 
These offsets are used to refine the template trajectory $\mathcal{T}^{\text{temp}_{c^S}}$ for the target supporting item.
The loss function for this head is a standard L2 loss, denoted as $L_{\mathcal{T}}$.

\noindent\textbf{Training Loss.} The total loss is the weighted sum of the three losses: $L_{\text{total}} = L_{\mathcal{T}} + 0.1 \times L_{a} + 0.1 \times L_{c}$.

\section{Experiment}
\label{sec:experiments}

% \begin{figure}[b!]
% % \centering
% % \hspace{-2mm}
%     \includegraphics[width=1.0\columnwidth,clip]{figures/Inference_Objects.pdf}
%     \caption{The objects and supporting items for evaluations.}
%     \label{fig:inference_objects}
% \vspace{-5mm}
% \end{figure}

% \begin{figure*}[t!]
% \centering
%     \includegraphics[width=0.9\textwidth]{figures/Real-World_Inference.pdf}
%     \caption{Our real-world setting and the visualization of the point clouds,  affordance maps, and the keypoint trajectories for the supporting items.} 
%     \label{fig:realworld_inference}
% \vspace{-5mm}
% \end{figure*}

\begin{figure*}[t!]
\centering
    \includegraphics[width=1.0\textwidth]{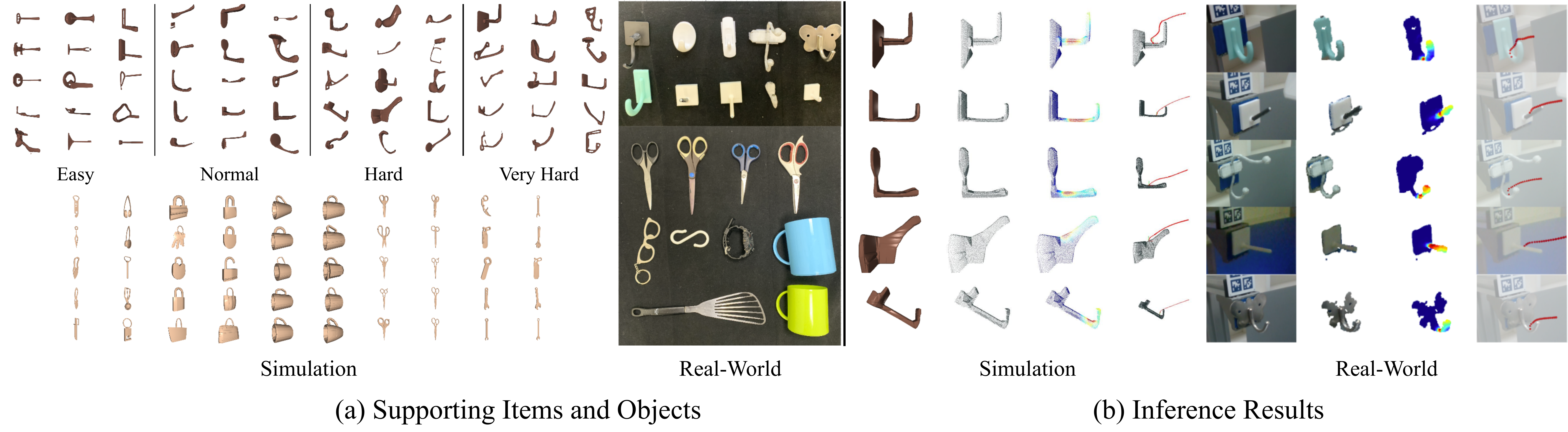}
    \caption{The supporting items and objects for evaluations and the visualization of point clouds, affordance maps, and the semantic keypoint trajectories for each supporting item.} 
    \label{fig:experiments}
\vspace{-5mm}
\end{figure*}

In this section, we present a series of experiments aimed at answering the following questions:
(1) Does our proposed method effectively generate suitable trajectories for hanging a diverse set of grasped objects onto novel supporting items with various shapes compared to existing methods?
%Can the proposed method effectively generate suitable trajectories for hanging a diverse set of grasped objects onto novel supporting items with various shapes? 
%
% adapted to the shape of the target supporting items and accommodate novel instances?
%
(2) Do shape-conditioned information and trajectory deformation play crucial roles in generating feasible actions for hanging?
(3) Does semantic keypoint trajectory (SKT) facilitate the hanging of diverse objects while being object-agnostic?
(4) Can our simulation-trained models transfer to real-world scenarios? 

\subsection{Experimental Setup}

The environment of a hanging task involves a supporting item and an object grasped by a robot arm. 
%
%In this task, t
The robot's objective is to determine a sequence of actions for hanging a grasped object onto a supporting item.
For evaluation, we assess the performance of all methods by measuring their success rates and inference times across 3,000 hanging processes involving 50 objects and 60 unseen supporting items (refer to Fig.~\ref{fig:experiments}(a)) in PyBullet~\cite{coumans:2021}.
These supporting items are manually categorized into four levels of difficulty: \textbf{Easy}, \textbf{Normal}, \textbf{Hard}, and \textbf{Very Hard} based on the complexity of the hanging part on them, with 15 assigned to each level.
%
% In \textbf{Easy}, the hanging part of each supporting item is designed to resemble a horizontal bar, requiring relatively straightforward actions for successful hanging. As the difficulty level increases, the hanging parts of the supporting items become more complex and challenging. 
% The evaluation process involved a total of 3000 hanging processes, with each method being thoroughly tested. 

\noindent\textbf{Baselines.}  We compare our method with four different baselines: kPAM 2.0~\cite{gao2021kpam2}, OmniHang~\cite{you2021omnihang}, VAT-Mart~\cite{wu2021vat}, and a modified version of VAT-Mart.
For kPAM 2.0, given the ground truth $SE(3)$ keypoint and the target hanging pose for each object, the robot moves the object from its initial pose to the target pose.
For OmniHang~\cite{you2021omnihang}, given the ground truth hanging pose for each object, we incorporate the collision estimation network they proposed into RRT-Connect~\cite{rrt-connect} to plan a trajectory for hanging.
%
%\ychen{you did not specify the actual hanging process of kPAM 2.0 and OmniHang.}
%
In the case of VAT-Mart and the modified version, given the $SE(3)$ keypoint for each object, we closely follow their implementation of the trajectory proposal network to generate 10 different SKTs for hanging each object onto a supporting item. 
The trial is considered successful if one of these trajectories successfully hangs the object onto the supporting item.  
%
% We modify VAT-Mart with the following two modifications. 
%
% First, we perform joint training of the affordance module and trajectory proposal module using the shared PointNet++ backbone. 
%
% Second, we utilize the shape-conditioned point features introduced in~\ref{methodology} instead of the contact point feature in VAT-Mart.
For the modified VAT-Mart, we utilize the shape-conditioned point features introduced in~\ref{subsec:method}, instead of the contact point feature used in VAT-Mart.
%information and the relevant geometry components essential for the hanging task. 
%
% The modification has the conditional variational auto-encoder trajectory proposal module, conditioned on the shape-conditioned point features.
%
% The design is similar to the proposed framework.
%
%obtained through max-pooling, corresponding to affordance scores above a certain threshold, similar to our proposed approach.
%
%Empirical observations demonstrate that these modifications significantly enhance the relationship between trajectory generation and the shape of the supporting item, resulting in a substantial improvement in success rate. 
%
For SCTDN, given the  $SE(3)$ keypoint for each object, we use one predicted SKT for evaluation.

% ================================================================
\subsection{Hanging Success Rate and Inference Time} 
% ================================================================
We report the success rate of robot hanging and the corresponding inference time in Table~\ref{table:main}. 
Our method demonstrates promising performance, achieving an average success rate of 83.7\%, 
%success rates 
compared to all baselines.
In particular, the proposed method shows superior performance in challenging cases, i.e., \textbf{Hard} and \textbf{Very Hard}.
%This performance is particularly notable when dealing with challenging supporting items, as we are able to maintain a success rate of nearly 80\%. 
%
% The modified VAT-Mart with the modeling of the task-relevant geometric structures of the supporting item  
The modified VAT-Mart with the shape-conditioned point features significantly outperforms the original VAT-Mart, especially in the \textbf{Very Hard} level. 
%in terms of execution success rates. 
%
% Specifically, in the \textbf{Very Hard} level, the success rate improves from 12.3\% to 62.4\%. 
%
Our experiments corroborate that the methods based on SKT outperform kPAM and OmniHang by a substantial margin.
%in terms of execution success rates.
%
% While OmniHang generates action sequences, the method faces challenges in detecting collisions under partial observability. 
While the success rate of OmniHang improves for challenging levels compared to kPAM, it still faces significant challenges in detecting collisions under partial observability.
%
% Additionally, due to non-smooth paths generated by RRT-Connect,
%'s inability to guarantee smooth paths poses difficulties for 
% the robot arm cannot successfully hang an object by following the action sequence. 
%in precisely moving the object to the specified position during execution.
%
% It is worth noting that, all learning-based methods are capable of completing inference (including the trajectory generation and hanging) within 20 milliseconds, while OmniHang, on average, requires 14.29 sec for planning. 
It is worth noting that, all learning-based methods are capable of generation trajectories within 20 milliseconds, while OmniHang, on average, requires 14.29 sec for planning. 
%
% This stark contrast makes Omnihang's planning time nearly ten thousand times longer than the inference time of learning-based methods.
% In other words, learning-based methods accelerate OmniHang by a factor of more than 1000 times. 

% ================================================================
\subsection{Ablation Study of the SCTDN Architecture} 
% ================================================================

In this experiment, we study the contribution of the shape-conditioned features and trajectory deformation module.
%on the robot's execution success rate within the SCTDN framework.
The results are shown in Table~\ref{tab:main_ablation}. 
The model achieves a 58.2\% success rate when we only align the retrieved template trajectory to the predicted contact point, without considering the deformation module.
We show that the deformation module boosts the hanging success rates significantly.
% %
% The results indicate the importance of the trajectory deformation head.
%
When we use the contact point feature for the deformation module, we observe a 15.3\% performance gain. 
On the other hand, a 25.5\% performance improvement is obtained with the shape-conditioned feature introduced in~\ref{subsec:method}.
The results indicate the importance of the trajectory deformation module and support that it is crucial to model the geometric structures of the supporting item's hanging part.

\begin{table}[!t]
% \hspace{0.03\linewidth}
% \begin{minipage}{.6\linewidth}
\centering
\caption{Ablation study of our network designs.}
\resizebox{1.0 \columnwidth}{!}{
\begin{tabular}
            {c @{\;} | c @{\;} | c @{\;} | c@{\;} | c}
            \toprule
            \multicolumn{1}{c}{Contact Point}  &
            \multicolumn{1}{c}{Contact Point Feature}  &
            \multicolumn{1}{c}{Part Feature}  &
            \multicolumn{1}{c}{Deformation}  &
            \multicolumn{1}{c}{Success Rate (\%)}
              \\
             \midrule
              \checkmark & & & &  58.2
              \\
              \checkmark & \checkmark & &  \checkmark & 73.5
              \\
              \checkmark & & \checkmark & \checkmark & 83.7
              \\
            \bottomrule
        
        \end{tabular}
}
\vspace{-6mm}
\label{tab:main_ablation}
\end{table}

% \begin{table*}[t!]
% % \begin{table}[t!]
% % \scriptsize
% \centering
% \caption{The hanging success rate of reference objects and other objects. \dag denotes our re-implementation of the baseline methods.}
% \resizebox{0.8\textwidth}{!}{
% % \resizebox{1.0\columnwidth}{!}{
%         \begin{tabular}
%             {@{} l @{\;} @{\;} | c @{\;} | c @{\;} c @{\;} c @{\;} c @{\;} c @{\;}}
%             \toprule
%             & 
%             \multicolumn{1}{c}{Reference Obj (\%)}  & 
%             \multicolumn{1}{c}{Mug (\%)}  & 
%             \multicolumn{1}{c}{Cooking Utensil (\%)}  & 
%             \multicolumn{1}{c}{Scissor (\%)}  & 
%             \multicolumn{1}{c}{Tool (\%)}  & 
%             \multicolumn{1}{c}{Others (\%)} 
%             % \\
%             % \midrule
%             %  Pose Trajectory & 86.7 & 65.2 & 48.0 & 18.7 & 14.7 & 74.8
%             \\
%              \midrule
%             %  VAT-Mart\dag~\cite{wu2021vat} & 68.3 & 51.6 & 49.3 & 44.9 & 49.7 & 47.8
%             % \\
%              Modified VAT-Mart\dag & 71.7 & 78.0 & 70.7 & 67.9 & 61.5 & 68.8
%             \\
%              \midrule
%               SCTDN (Ours) & \textbf{91.7} & \textbf{88.8} & \textbf{81.2} & \textbf{75.3} & \textbf{78.0} & \textbf{84.3}
%             \\
%             \bottomrule
%         \end{tabular}
% }
% \label{tab:skt_pose}
% \vspace{-3mm}
% % \end{table}
% \end{table*}

\subsection{One Semantic Keypoint Trajectory for Various Objects}
% To answer the first question, we conduct evaluations using two types of trajectories: pose trajectory and semantic keypoint trajectory. 
In this experiment, we assess whether an SKT can be applied to various objects for each supporting item.
We deploy modified VAT-Mart and SCTDN to predict SKT for each supporting item.
% semantic keypoint trajectories effectively facilitate the hanging of various objects for each supporting item.
%
We compare the overall success rates of hanging a reference object and five additional object categories (each with ten instances) onto 60 different supporting items
%, \textcolor{blue}{each associated with a single predicted SKT.}
%
Note that the reference object is used to generate trajectories for all supporting items, discussed in Sec.~\ref{subsec:data_collection}.
%served as the reference for
%, acting as a canonical trajectory. 
%
% The pose trajectory for each supporting item is obtained using the RRT-Connect~\cite{rrt-connect} method, assuming complete knowledge of the shape for collision checking. 
%
% This ensured that the pose trajectories is collision-free for the reference object.
%
% Three methods were employed to obtain the semantic keypoint trajectory, i.e., VAT-Mart, modified VAT-Mart, and the proposed method. 
%Two methods were employed to obtain the SKT, i.e., modified VAT-Mart, and our proposed method. 
%
%\textcolor{blue}{They predict a single SKT for each supporting item to hang objects.}
%
%The success rates for each method are presented in Table~\ref{tab:skt_pose}.
%The results are reported in Table~\ref{tab:skt_pose}.
%
The results reported in Table~\ref{tab:skt_pose} indicate that SKT enables similar success rates of hanging different objects compared to the reference object.
This greatly enhances the versatility and applicability of SKT to a wide range of everyday objects.
Moreover, the SCTDN demonstrates a promising capability to generate SKT, compared to modified VAT-Mart.
%
%different types of objects to achieve success rates close to that of the reference object. 
%
% In contrast, the use of the pose trajectory method results in significant variations in success rates because the representation is not object-agnostic. 
%
% The exception is when objects' shapes are similar to the reference such as the lock shapes in Fig~\ref{fig:inference_objects}. 

% ================================================================
\subsection{Real-World Evaluation}
% ================================================================

We conducted real-world experiments to evaluate the transferability of our simulation-trained models to real-world scenarios.
Our experimental setup included a Franka Emika robot arm, an Intel D-435 camera, an object grasped by the robot arm, and a supporting item. 
In the real-world evaluation, we used a total of 10 objects and 10 supporting items, as shown in Fig.~\ref{fig:experiments}(a).
To handle noise in real-world point clouds, we defined the foreground region of the supporting item using an AprilTag and applied DBSCAN~\cite{ester1996density} to remove noisy points. 
Alternatively, one can use image segmentation models~\cite{kirillov2023segment} to obtain a segmented point cloud.

The hanging success rate evaluation involved the robot hanging all 10 objects onto the 10 supporting items, resulting in 100 hanging processes. 
The overall success rate for hanging was 92\%, indicating the effectiveness of our approach in guiding various objects onto different supporting items in real-world scenarios.
The right part of Fig.~\ref{fig:experiments}(b) visualizes the predicted Semantic Keypoint Trajectories (SKTs) in simulation and real-world settings. 
%real point clouds. 
%
For additional qualitative results and demonstrations, please refer to our video submission and the \href{https://chialiang86.github.io/skt-hang.github.io/}{\textbf{webpage}}.
%for more detailed information.
% As depicted in Fig.\ref{fig:realworld_setting}(a), our experimental setup included a Franka Emika robot arm, an Intel D-435 camera mounted on a movable stand, and a designated supporting item placed on another movable stand. 
% Our experimental setup included a Franka Emika robot arm, an Intel D-435 camera, an object grasped by the robot arm, and a supporting item. 
% %
% We used a total of 10 objects and 10 supporting items in the real world for our hanging evaluation as depicted in Fig.~\ref{fig:experiments}(a). %, which can be seen in.
% %
% % To capture the point cloud of the supporting item, we utilized an Intel D-435 camera. 
% %
% To reduce background noise from real-world point clouds, we define the foreground region of the supporting item using an Apriltag and employ DBSCAN~\cite{ester1996density} to remove noisy points.
% %
% Alternatively, one can also use image segmentation models~\cite{kirillov2023segment} to obtain a segmented point cloud. 

% To evaluate the hanging success rate, the robot hung all 10 objects onto the 10 supporting items, resulting in 100 hanging processes. 
% %
% The overall success rate for hanging was 92\%, signifying that our approach effectively guides diverse objects onto various supporting items, even in real-world scenarios.
% %
% The right part of Fig.~\ref{fig:experiments}(b) visualize the SKTs for using real point clouds.
% %
% For additional qualitative results and demos, please refer to our project page for more detailed information.

\subsection{More Ablation Studies of SCTDN and VAT-Mart}
We conduct additional ablation studies to compare the hanging success rates of SCTDN and VAT-Mart-based methods across various trajectory lengths $T\in\{10, 20, 40\}$ and different waypoint dimensions (position-only and position-and-rotation waypoints) within the SKT. 
For comprehensive discussions and detailed table information, please visit our project webpage.

\begin{table}[t!]
% \begin{table}[t!]
% \scriptsize
\centering
\caption{The hanging success rate of reference objects and other objects. \dag~denotes our re-implementation of the baseline methods.}
% \resizebox{0.8\textwidth}{!}{
\resizebox{0.8\columnwidth}{!}{
        \begin{tabular}
            {@{} l @{\;}  @{\;} | c @{\;} | c @{\;}}
            \toprule
            & 
            \multicolumn{1}{c}{Modified VAT-Mart \dag}  & 
            \multicolumn{1}{c}{SCTDN (Ours)}
            \\
             \midrule
             Reference Object (\%)  & 71.7 & 91.7
            \\
             \midrule
             Mug (\%)  & 78.0 & 88.8
            \\
             Cooking Utensil (\%)  &  70.7 & 81.2
            \\
             Scissor (\%)  & 67.9 & 75.3
            \\
             Tool (\%)  & 61.5 & 78.0
            \\
             Others (\%)  & 68.8 & 84.3
            \\
            \bottomrule
        \end{tabular}
}
\label{tab:skt_pose}
\vspace{-5mm}
% \end{table}
\end{table}

% \begin{figure}[t!]
% \centering
%     \includegraphics[1.0 \columnwidth]{figures/Real-World_Objects.pdf}
%     \caption{The real-world objects and supporting items for evaluation.} 
%     \label{fig:realworld_objects}
% % \vspace{-5mm}
% \end{figure}

% \begin{figure}[t!]
% \centering
%     \includegraphics[1.0 \columnwidth]{figures/Real-World_Inference.pdf}
%     \caption{The visualization results of the real-world point clouds, the predicted affordance maps, and the semantic keypoint trajectories for each supporting item.} 
%     \label{fig:realworld_inference}
% % \vspace{-5mm}
% \end{figure}

\section{Conclusion}
\label{sec:conclusion}

In this paper, we propose \textit{Semantic Keypoint Trajectory}, a novel representation for hanging everyday objects. 
This representation is actionable and interpretable, specifying both where and how to hang while being object-agnostic. This enables the robot to efficiently learn the complex hanging task. 
To predict such a representation, we introduce a shape-conditioned trajectory deformation network by leveraging the insight that crucial geometric parts play a significant role in trajectory generation. 
By deforming existing trajectories, we are able to obtain a semantic keypoint trajectory adapted to the shape of novel supporting items.
We conducted empirical evaluations through robot executions to hang various objects on a diverse range of supporting items, demonstrating the effectiveness of our proposed approach in both the simulation and the real world.
Our work holds potential for broader application in other relevant tasks.

% Although semantic keypoint trajectory demonstrates the ability to be usable for other objects, there is still room for improvement in terms of its execution success rate. 
% %
% Therefore, in future work, we will explore how to utilize semantic keypoint trajectory as an initial guess for the task and make fine adjustments based on the characteristics of each object to enhance the success rate. 
% %
% Additionally, we observed that the execution success rate for easy supporting items is not the highest in the success rate evaluation. 
% %
% Our analysis suggests that this is mainly due to the simplicity of the shapes in easy cases, which leads to a higher fault tolerance of the semantic keypoint trajectory towards the pivot object during data collection. 
% %
% As a result, its reusability is compromised, resulting in suboptimal training outcomes. 
% %
% To address this issue, we will improve the data collection pipeline to better showcase the advantages of semantic keypoint trajectory.

\clearpage
\bibliographystyle{IEEEtran}
\bibliography{egbib}

% \section{Conclusion}
% \input{Content/7-conclusion}

\end{document}